\documentclass{article}
\usepackage{times}
\usepackage{iclr2027_conference}

\usepackage{amsmath,amsfonts,bm}

\def\eqref#1{equation~\ref{#1}}

\def\1{\bm{1}}

\DeclareMathAlphabet{\mathsfit}{\encodingdefault}{\sfdefault}{m}{sl}
\SetMathAlphabet{\mathsfit}{bold}{\encodingdefault}{\sfdefault}{bx}{n}

\usepackage[hidelinks]{hyperref}
\usepackage{url}
\usepackage{tikz}
\usetikzlibrary{arrows.meta,positioning,shapes.geometric, calc, fit, backgrounds}
\usepackage{booktabs}
\usepackage{siunitx}
\usepackage{subcaption}

\newcommand{\pmpar}[1]{\,(#1)}

\title{Predictive Dual Smoothing \\for Column Generation}

\author{Senne Berden \\
KU Leuven
\And
Noah Schutte \\
TU Delft
\And
Andrea Lodi \\
Cornell Tech
\And
Tias Guns \\
KU Leuven
}

\iclrfinalcopy

\begin{document}

\maketitle

\makeatletter
\lhead{}
\makeatother

\begin{abstract}

Solving large-scale linear programs efficiently is an important challenge in many optimization settings. A key technique is column generation, which alternates between solving the master problem over a restricted subset of the variables, and using a pricing subproblem to identify new variables to add. The pricing subproblem is guided by the dual solution of the current restricted master problem, but oscillations in these dual solutions can substantially slow convergence. Dual stabilization methods address this issue. Dual smoothing is a common stabilization method, which guides the pricing subproblem using a combination of the current dual solution and duals from previous iterations. However, while past dual solutions can stabilize the dual trajectory, they do not necessarily guide pricing towards useful new variables. We therefore introduce \textit{predictive dual smoothing}, which instead combines the current dual solution with a learned prediction of future duals to steer pricing towards variables that are more useful in subsequent iterations. The predictor is trained offline using supervision extracted from standard column generation trajectories and is used only to modify the pricing subproblem's objective function, while exact reduced-cost checks and fallback pricing with the unsmoothed duals preserve correctness. Experiments on cutting stock and generalized assignment problems show that predictive dual smoothing substantially reduces generated columns and wall-clock time relative to standard column generation and existing classical and learned stabilization methods. These gains extend to out-of-distribution instance sizes, and predictive smoothing provides further improvements when combined with strong classical stabilization.
\end{abstract}

\section{Introduction}

Column generation (CG) is a widely used technique for solving large-scale linear programs whose full set of variables is too large to enumerate explicitly. Examples include cutting stock, routing and scheduling problems \citep{lubbecke2005selected, desaulniers2006column}. Rather than solving the full problem directly, CG alternates between a restricted master problem (RMP), containing only a subset of decision variables (and thus \textit{columns} in the constraint matrix), and a pricing subproblem, hereafter simply called the pricing problem, that searches for new columns (i.e., variables) whose addition would improve the RMP's objective value. The pricing problem is guided by the dual solution of the current RMP, which determines which columns are attractive to generate next.

The duals have an economic interpretation that explains why they are used for pricing. Each dual value is a `shadow price' associated with a constraint. It expresses the marginal improvement in the optimal RMP objective obtained if that constraint were relaxed, and thus reflects how costly that constraint is to satisfy using the columns currently available in the RMP. At every iteration $t$, the pricing problem takes optimal RMP duals \(\boldsymbol{\pi}_t\) as input and returns the column that best addresses the needs of the RMP expressed by these duals, by favoring columns that make currently expensive constraints easier to satisfy. From the dual perspective, each column that is not yet in the RMP corresponds to a constraint that is absent from the current dual problem. Pricing identifies the missing column whose corresponding dual constraint is most violated by the current dual solution.

However, the current duals reflect only the current incomplete column set. A dual value may be high only because the columns available so far happen to address the corresponding constraint poorly. As other columns are added, the duals can therefore change substantially from one iteration to the next. In practice, this can lead to \textit{dual oscillation}, which can cause pricing to repeatedly generate columns that are attractive only under temporary dual prices, unnecessarily enlarging the RMP and slowing convergence to the optimal solution~\citep{desrosiers2005primer}. This has motivated research into \textit{dual stabilization} techniques~\citep{du1999stabilized}.

A common stabilization strategy is \emph{dual smoothing}~\citep{wentges1997weighted, neame2000nonsmooth, pessoa2018automation}. Instead of pricing directly with the current RMP dual \(\boldsymbol{\pi}_t\), smoothing combines it with a more stable reference point,
$$
    \widetilde{\boldsymbol{\pi}}_t
    =
    (1-\alpha)\boldsymbol{\pi}_t
    +
    \alpha\boldsymbol{h}_t,
$$
where $\alpha$ is a hyperparameter controlling the strength of smoothing, and $\boldsymbol{h}_t$ is a reference point constructed from dual values observed in previous CG iterations.
This can reduce the sensitivity of pricing to dual oscillations, and in turn improve convergence. However, the duals incorporated through the reference point come from earlier, smaller RMPs, and the needs they express may already have been addressed by columns generated in the meantime. As a result, the reference point can continue to direct pricing towards needs that have already been resolved. Thus, smoothing may reduce oscillation, but does not necessarily direct pricing towards columns that will remain useful as the RMP evolves.

This suggests a different choice of reference point. If the duals that will be encountered later in the CG trajectory were known in advance, they could be used as a forward-looking reference point to smooth the current duals. The resulting pricing signal would still reflect the needs of the current RMP, while also anticipating needs that would emerge in the following iterations. This would both counteract the sensitivity of the pricing problem to short-term dual oscillations and favor columns that remain useful as the RMP evolves. This could accelerate CG convergence significantly. However, when the pricing problem must be solved, future duals are, of course, not yet available.

Our key idea is therefore to predict these future duals from the current CG state and use the predictions as smoothing reference points. We call this \emph{predictive dual smoothing}. Given the current CG state $s_t$, we predict each component of the dual solution $k$ RMP updates ahead using a shared predictor $m_\theta$, and collect these predictions in $\widehat{\boldsymbol{\pi}}_{t+k}$. Pricing is then performed
using smoothed duals
\[
    \widetilde{\boldsymbol{\pi}}_t^{(k)}
    =
    (1-\alpha)\boldsymbol{\pi}_t
    +
    \alpha\widehat{\boldsymbol{\pi}}_{t+k}.
\]
The prediction horizon $k$ determines how far ahead the smoothing reference looks, while $\alpha$ controls how strongly the prediction influences pricing. The prediction is used only to guide the search for new columns. Any generated column is verified against the current RMP solution, and standard pricing (i.e., with $\alpha = 0$) is used whenever the smoothed pricing step fails to identify a valid improving column. Learning can therefore change the sequence of generated columns without affecting the correctness or optimality guarantees of CG.

We train the future-dual predictor offline from trajectories generated by standard CG on training instances from the target problem class. Each visited state $s_t$ can be paired directly with the dual solution observed at iteration $t+k$, so a single CG trajectory provides many supervised training pairs. Moreover, the same recorded trajectory can be reused for different prediction horizons (e.g., when tuning $k$) by changing only the target iteration.

Our contributions are:
\begin{itemize}
    \item We introduce \emph{predictive dual smoothing}, which uses predictions of future RMP dual states as a reference point for smoothed pricing, with the prediction horizon controlling how far ahead the method looks.
    \item We show that future duals can be learned using simple supervised learning from ordinary CG trajectories, with each trajectory providing many training pairs and reusable supervision across prediction horizons.
    \item We extensively evaluate predictive dual smoothing on cutting stock and generalized assignment problems, showing substantial reductions in generated columns and runtime over standard CG and existing classical and learned stabilization methods. Our experiments further characterize the effects of prediction horizon, smoothing strength, and training-data size. They also study both in-distribution and out-of-distribution generalization across instance scales, and show that predictive smoothing can even provide additional gains when combined with strong classical stabilization methods.

\end{itemize}

\section{Related Work}

The integration of machine learning into CG has received increasing attention in recent years, with learning-based components becoming involved in several parts of the CG loop.
A first line of work concerns \textit{column selection}, which first generates a \textit{pool} of solutions to the pricing problem, and subsequently uses a learned classifier to decide which of these columns to add to the RMP. \citet{morabit2021machine} use imitation learning to train the classifier, with supervision obtained through a relatively expensive one-step lookahead expert. Subsequent work instead uses reinforcement learning to train the classifier, with the goal of optimizing single-column choices for long-term CG performance rather than immediate improvement \citep{chi2022deep}. Later extensions generalize this approach to selecting multiple columns per iteration, either with a fixed \citep{multiple_columns} or variable \citep{hu2025ffcg} number of columns.
Other approaches intervene inside or around pricing: \citet{shen2022pricing} predict pricing solutions for graph coloring, \citet{ML_arc_selection} learn to restrict the network used by routing pricing problems, and \citet{ml_branch_and_price} consider settings with multiple pricing problems per CG iteration and learn which one to solve first.
Other directions include learning to \textit{remove} redundant columns \citep{fang2023accelerating}, and predicting whether sampled columns are likely to appear in an optimal integer solution \citep{sun2022learning}.

Most relevant to our work are methods that use learning to influence the duals guiding CG. \citet{babaki2022coil} address non-uniqueness of RMP dual solutions and learn, via imitation learning and a differentiable optimization layer, which point on the current dual-optimal face to use for pricing. \citet{kraul2023machine} instead predict the optimal full-master dual values for the cutting stock problem and use these predictions as stabilization centers in a conventional box-stabilized CG scheme. Similarly, \citet{shen2024adaptive} predict the eventual optimal dual solution and use it to guide adaptive stabilization for graph coloring. More recently, \citet{fang2025learning} use reinforcement learning to output stabilized dual vectors directly for pricing.

Our approach differs from these methods in both the prediction target and how the prediction is used. Rather than selecting among current RMP-optimal duals or predicting the eventual full-master optimum, we predict dual solutions at a finite horizon along the future CG trajectory. These predictions are used directly as smoothing reference points for pricing, rather than as centers of a box-stabilized master. This not only reduces the sensitivity of pricing to short-term dual oscillations, but also biases pricing towards columns that address needs anticipated to arise in subsequent CG iterations. Finally, our predictor uses generic features of the CG state rather than problem-specific representations, allowing predictive smoothing to be applied across different problem classes.

\section{Preliminaries: Column Generation}
\label{sec:preliminaries}

CG solves large linear programs without explicitly enumerating all columns. It maintains an RMP over a subset of columns and uses its dual solution to identify improving columns through a pricing problem.

\paragraph{Illustrative example: cutting stock.}
We use the cutting stock problem (CSP) as an illustrative example. In the CSP, we are given an unlimited supply of stock rolls of fixed length $L$ and a collection of item types that must be cut from these rolls. Item type $i\in\{1,\ldots,n\}$ has length $l_i$ and
demand $d_i$. The goal is to satisfy all demands while using as few stock
rolls as possible.

A single stock roll can be cut in many different ways, called \textit{cutting patterns}. For example, one cutting pattern may contain two items of one type and three of another, while a different pattern may contain another combination. We represent pattern $p$
by $x_p=(x_{1p},\ldots,x_{np})^\top\in\mathbb{N}^n,$
where $x_{ip}$ denotes the number of items of type $i$ produced by pattern
$p$. A pattern is feasible if the total length of the items assigned to the
roll does not exceed $L$. The set of all feasible patterns is thus $
    \mathcal{P}
    =
    \left\{
        x\in\mathbb{N}^n:
        \sum_{i=1}^n l_i x_i\leq L
    \right\}
$.

Each feasible pattern defines a master variable \(\lambda_p\), denoting the number of stock rolls cut according to pattern \(p\). As is standard in CG, we work with the LP relaxation of the
resulting formulation, so $\lambda_p \in \mathbb{R}_+$.\footnote{In an exact integer-programming
method, the same CG procedure can be used to solve LP relaxations within a
branch-and-price framework.}
The Gilmore-Gomory formulation \citep{gilmore1961linear} is then
\begin{subequations}
\label{eq:csp_master}
\begin{align}
    \min_{\lambda\geq0}\quad&
        \sum_{p\in\mathcal{P}}\lambda_p
        \\
    \text{s.t.}\quad&
        \sum_{p\in\mathcal{P}}x_{ip}\lambda_p
        \geq d_i,
        \qquad i=1,\ldots,n.
\end{align}
\end{subequations}
The objective minimizes the number of stock rolls used, while each constraint $i$ ensures that enough items of type $i$ are produced to satisfy demand $d_i$.

The challenge is that the number of feasible cutting patterns can be enormous. Rather than include all of them at once, CG maintains only a subset $\mathcal{P}_t\subseteq\mathcal{P}$ at iteration $t$ and solves the master problem using only the corresponding variables
$\{\lambda_p:p\in\mathcal{P}_t\}$. This restricted problem is the RMP.

Solving the RMP produces not only a primal solution but also a dual solution $\boldsymbol{\pi}_t\in\mathbb{R}_+^n$, with one dual value associated
with each demand constraint. These dual values define the objective of the \emph{pricing problem}, which is used to generate new columns. It searches over all feasible cutting patterns for a column that could improve the current RMP solution.

For any pattern $p$, its \textit{reduced cost} under the current RMP dual is $\bar c_p(\boldsymbol{\pi}_t) = 1-\boldsymbol{\pi}_t^\top x_p$, where $1$ is the cost of using one stock roll and $\boldsymbol{\pi}_t^\top x_p$ is the total dual value of the items produced by the pattern. Thus, the reduced cost measures the pattern’s cost relative to the value it provides under the current dual prices.
Adding a pattern to the RMP can improve its objective whenever the pattern's reduced cost is negative. Equivalently, a pattern with negative reduced cost corresponds to a currently violated constraint in the dual of (\ref{eq:csp_master}). When CG generates a new pattern, it finds the pattern with the most negative reduced cost by solving
\begin{equation}
    x_t^\star
    \in
    \arg\max_{x\in\mathbb{N}^n}
    \left\{
        \boldsymbol{\pi}_t^\top x:
        \sum_{i=1}^n l_i x_i\leq L
    \right\}.
    \label{eq:csp_pricing}
\end{equation}

If $1-\boldsymbol{\pi}_t^\top x_t^\star<0$, the corresponding pattern is added to the RMP and the RMP is solved again. Otherwise, no feasible pattern has negative reduced cost, which certifies that the current RMP solution is optimal for problem~(\ref{eq:csp_master}), and CG is terminated. We refer to this procedure, with exact pricing and one column of minimum reduced cost added per iteration, as \emph{standard CG}.

\paragraph{Dual stabilization and smoothing.}
\label{sec:dual_stabilization}

Standard CG uses the current RMP dual $\boldsymbol{\pi}_t$ directly to define the pricing objective. However, this dual only reflects the needs of the \textit{current} restricted column set. Some of these needs may disappear once new columns are added. The dual solution can therefore change substantially after each RMP update, causing pricing to repeatedly target temporary needs and generate columns that remain useful only for a small number of iterations. This behavior can lead to slow convergence and has motivated research into \textit{dual stabilization} methods.

Stabilization can intervene at different points in the CG loop. Some stabilization methods modify the RMP so that its dual solution is encouraged to remain near a stability center, as in the box/penalty approach of \citet{du1999stabilized}. In contrast, \textit{dual smoothing} leaves the RMP unchanged and instead modifies the dual vector used in pricing. These mechanisms are complementary: smoothing can be applied either to the dual of an ordinary RMP or to the dual produced by a stabilized master.

In dual smoothing, the dual in the pricing objective is replaced by a convex combination
\begin{equation}
    \widetilde{\boldsymbol{\pi}}_t
    =
    (1-\alpha)\boldsymbol{\pi}_t
    +
    \alpha \boldsymbol{h}_t,
    \qquad
    \alpha\in[0,1],
    \label{eq:generic_smoothing}
\end{equation}
where $\boldsymbol{h}_t$ is a reference point and $\alpha$ controls the
strength of the stabilization.

Different smoothing methods differ primarily in how the reference point $\boldsymbol{h}_t$ is constructed. For example, Neame smoothing \citep{neame2000nonsmooth} uses the previous smoothed
pricing vector $\boldsymbol{h}_t=\widetilde{\boldsymbol{\pi}}_{t-1}$, whereas Wentges smoothing \citep{wentges1997weighted} uses an incumbent dual vector associated with the best Lagrangian bound found so far.

In both cases, the reference point is constructed from duals obtained at earlier stages of CG. As columns are added, however, the needs reflected in these earlier duals may already have been addressed. The resulting pricing signal can therefore remain influenced by needs that are no longer important for the current RMP. While this can reduce the sensitivity of pricing to dual oscillations, it does not necessarily favor columns that will remain useful as the RMP continues to evolve.

\section{Predictive Dual Smoothing}
\label{sec:predictive_smoothing}

Conventional dual smoothing constructs its reference point from information observed in previous CG iterations. Predictive dual smoothing instead uses a prediction of a future dual state, with the goal of guiding pricing towards columns that remain useful as the RMP evolves. Our approach is illustrated in Figure~\ref{fig:predictive_smoothing_pipeline}. We now discuss its components in turn.

\definecolor{accent}{RGB}{61,90,254}
\definecolor{accentbg}{RGB}{224,229,255}
\definecolor{inkc}{RGB}{45,45,52}
\definecolor{edgec}{RGB}{158,160,170}
\definecolor{procbg}{RGB}{243,243,246}
\definecolor{decbg}{RGB}{255,229,143}
\definecolor{decln}{RGB}{219,150,10}
\begin{figure}
\centering
\resizebox{0.9\linewidth}{!}{%
\begin{tikzpicture}[
>={Latex[length=2.2mm,width=1.8mm,round]},
font=\sffamily\small,
node distance=0.44cm and 0.38cm,
proc/.style={
rounded corners=3pt, fill=procbg, draw=inkc, line width=0.6pt,
text=inkc, minimum height=1.15cm, text width=2.05cm,
align=center, inner sep=3pt
},
learned/.style={
proc, fill=accentbg, draw=accent, line width=0.8pt, text=inkc
},
dec/.style={
diamond, aspect=2.4, fill=decbg, draw=decln, line width=1.3pt,
text=inkc, inner sep=0pt, text width=1.65cm,
align=center, font=\sffamily\bfseries\footnotesize
},
arr/.style={->, draw=inkc, line width=0.7pt},
pill/.style={
font=\sffamily\footnotesize, text=inkc,
fill=white, inner xsep=2.5pt, inner ysep=1pt, rounded corners=1.5pt
}
]
\node[proc] (rmp) {Solve RMP\\[1pt] $\boldsymbol{\pi}_t$};
\node[learned, right=of rmp] (pred)
{Predict\\[1pt] $\widehat{\boldsymbol{\pi}}_{t+k}$};
\node[learned, right=of pred] (blend)
{Smooth\\[1pt] $\widetilde{\boldsymbol{\pi}}_t^{(k)}$};
\node[proc, right=of blend] (price)
{Price with\\[1pt] $\widetilde{\boldsymbol{\pi}}_t^{(k)}$};
\node[dec, right=of price] (chk1) {$\bar c(\widehat x_t)<0$?};
\begin{scope}[on background layer]
\node[draw=accent!100, dashed, dash pattern=on 2pt off 1.6pt,
rounded corners=5pt, line width=0.7pt,
fit=(pred)(blend), inner sep=5.5pt] (fwdbox) {};
\end{scope}
\node[font=\sffamily\bfseries\footnotesize, text=accent, above=2pt of fwdbox]
{Predictive dual smoothing};
\draw[arr] (rmp) -- (pred);
\draw[arr] (pred) -- (blend);
\draw[arr] (blend) -- (price);
\draw[arr] (price) -- (chk1);
\node[proc, below=0.78cm of chk1] (fb)
{Fallback:\\[1pt] price with $\boldsymbol{\pi}_t$};
\node[dec, left=0.42cm of fb] (chk2) {$\bar c(x^{\mathrm{fb}}_t)<0$?};
\node[proc, left=1.15cm of chk2] (term) {Terminate\\[1pt] (LP optimal)};
\draw[arr] (chk1) -- node[pill] (nolbl) {no} (fb);
\draw[arr] (fb) -- (chk2);
\draw[arr] (chk2) -- node[pill,midway] {no} (term);
\coordinate (bus)  at ($(chk2.south)+(0,-0.60)$);
\coordinate (busL) at (rmp.south |- bus);
\coordinate (busR) at ($(chk1.east)+(0.35,0)$);
\draw[draw=inkc, line width=0.7pt] (chk1.east) -- (busR);
\draw[draw=inkc, line width=0.7pt]
(busR) -- node[pill] {yes} (busR |- bus);
\draw[draw=inkc, line width=0.7pt]
(chk2.south) -- node[pill] {yes} (chk2.south |- bus);
\draw[arr] (busR |- bus) -- (busL) -- (rmp.south);
\node[pill, text=inkc] at ($(busL)!0.42!(chk2.south |- bus)$)
{Add column};
\end{tikzpicture}
}
\caption{
Predictive dual smoothing.
The learned predictor and the smoothing step (blue) replace conventional dual smoothing. A column proposed under $\widetilde{\boldsymbol{\pi}}_t^{(k)}$ is accepted only if its reduced cost is negative under the true current dual $\boldsymbol{\pi}_t$. When the test fails, standard pricing with $\boldsymbol{\pi}_t$ either supplies an improving column or certifies LP optimality.
}
\label{fig:predictive_smoothing_pipeline}
\end{figure}
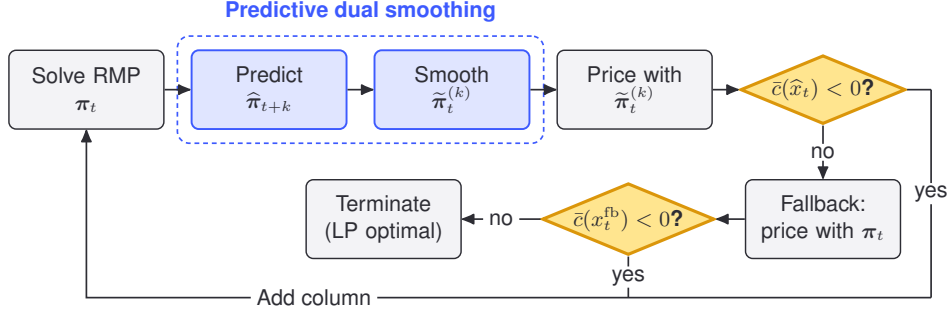

Let $\boldsymbol{\pi}_t$ denote the dual vector given to pricing by the underlying CG procedure. For standard CG, this is the current RMP dual. When a stabilized RMP is used, it is the corresponding dual of this stabilized RMP. Let \(s_t\) denote the CG state at iteration \(t\), comprising information from the current RMP and its solution, the pricing formulation, and the progress of the CG procedure. Given \(s_t\), we predict the dual solution \(k\) RMP updates ahead and denote the resulting vector by \(\widehat{\boldsymbol{\pi}}_{t+k}\). We then define the smoothed pricing vector as
\begin{equation}
    \widetilde{\boldsymbol{\pi}}_t^{(k)}
    =
    (1-\alpha_t)\boldsymbol{\pi}_t
    +
    \alpha_t\widehat{\boldsymbol{\pi}}_{t+k},
    \label{eq:predictive_smoothing}
\end{equation}
where $\alpha_t\in[0,1]$ controls the influence of the predicted future dual and is initialized to a hyperparameter $\alpha_0$. Standard pricing is recovered for $\alpha_t=0$, whereas $\alpha_t=1$ prices directly using the predicted future duals. Horizon $k$ controls how far ahead the reference point lies along the CG trajectory: $k=1$ targets the next RMP duals, while larger $k$ targets more distant future states. We also consider the terminal duals $\boldsymbol{\pi}_T$, to support smoothing with respect to a prediction of the final duals.

\subsection{Preserving Correctness}
By using smoothed objective $\widetilde{\boldsymbol{\pi}}_t^{(k)}$ in pricing, we lose the guarantee that pricing will produce a column with negative reduced cost whenever one exists. Without an additional safeguard, this would break the correctness and termination guarantees of CG. We therefore first evaluate whether the actual reduced cost of the produced column is negative, using the current duals $\boldsymbol{\pi}_t$.

If this test passes, the column is added to the RMP, and CG continues. If this test fails, we resort to standard pricing with $\boldsymbol{\pi}_t$ as a fallback. If this produces a column with negative reduced cost, this column is added to the RMP. Otherwise, the fallback has certified that no column with negative reduced cost exists, and CG terminates.
Predictive smoothing can therefore change \textit{which} improving columns are generated and may occasionally require an additional pricing call, but it cannot add a non-improving column or cause premature termination.

Whenever a fallback occurs, we reduce the smoothing strength as $\alpha_{t+1}=\gamma\alpha_t$, where $\gamma\in[0,1]$. Otherwise, $\alpha_{t+1}=\alpha_t$. Since predictive pricing becomes more likely to fail as CG approaches convergence, this decay progressively moves pricing towards the current dual and avoids repeated fallbacks. We ablate this mechanism, and evidence its importance, in Appendix~\ref{app:gamma_ablation}.

\subsection{Learning Future Duals}
\label{sec:learning_future_duals}

We train the future-dual predictor offline from CG trajectories generated on a set of training instances. For each instance, we run the base CG procedure and record the sequence of states $s_0,\ldots,s_T$ and corresponding dual solutions $\boldsymbol{\pi}_0,\ldots,\boldsymbol{\pi}_T$, where $T$ denotes the terminal RMP iteration. The base procedure may use either a standard or stabilized RMP.

We formulate future-dual prediction at the level of individual RMP constraints. For each constraint $i$ in state $s_t$, we construct a fixed-dimensional feature vector $f_i(s_t)$ and predict the corresponding future dual value with a shared model $m_\theta$. For prediction horizon $k$, the target is $\pi_{i,\tau(t,k)}$, where $\tau(t,k)=\min\{t+k,T\}$. States within $k$ iterations of convergence use the terminal dual as their target rather than being discarded. Let $\mathcal D_k$ denote the resulting set of constraint-level training examples. We train $m_\theta$ by minimizing
\begin{equation}
    \mathcal{L}(\theta)
    =
    \frac{1}{|\mathcal D_k|}
    \sum_{(t,i)\in\mathcal D_k}
    \left(
        m_\theta(f_i(s_t))
        -
        \pi_{i,\tau(t,k)}
    \right)^2.
    \label{eq:future_dual_loss}
\end{equation}
A single collection of trajectories can be reused across prediction horizons. Changing $k$ only changes which future dual supplies the target label. Further training details are given in Appendix~\ref{app:learning_details}.

\paragraph{Feature representation.}
For each dual value $\pi_i$, we construct a fixed-dimensional feature vector $f_i(s_t)$ from quantities already available in the current RMP and pricing formulation. The features are designed to capture four aspects of the CG state relevant to predicting how the dual associated with constraint $i$ will evolve: (i) the \emph{global CG state}, such as the iteration, RMP size, and objective value; (ii) the \emph{constraint state}, including the current dual value, right-hand side, and primal activity of constraint $i$; (iii) the \emph{pricing structure}, summarizing how variables associated with constraint $i$ enter the pricing objective and constraints; and (iv) the \emph{column context}, summarizing how constraint $i$ is represented among the columns already present in the RMP. All features are expressed through generic CG quantities rather than problem-specific representations, allowing the same feature construction to be used across problem classes. Full feature definitions are given in Appendix~\ref{app:features}.

\paragraph{Predictor.}
Because the predictor is evaluated at every CG iteration, its inference cost directly offsets any runtime gains obtained from better pricing. We therefore deliberately use a lightweight multilayer perceptron (MLP) $m_\theta$. This MLP is single-output, and predicts one future dual value at a time, i.e., $\widehat{\pi}_{i,t+k} = m_\theta(f_i(s_t))$. These evaluations are independent and can be batched in practice.

This approach is permutation equivariant: reordering the RMP constraints simply reorders the corresponding predictions, without changing their values. The predictor therefore does not depend on the arbitrary ordering of the RMP constraints. Moreover, because its input and output dimensions are fixed for a single constraint, the same network can be applied to instances with different numbers of RMP constraints, without modifying the architecture.

\section{Experiments}
\label{sec:experiments}

We evaluate predictive dual smoothing on two structurally different CG problems: the cutting stock problem and the generalized assignment problem (GAP). We investigate its effectiveness relative to standard CG and existing stabilization methods, its sensitivity to the prediction horizon, smoothing strength, and amount of training data, and its ability to generalize across instance sizes and complement strong classical stabilization.

\paragraph{Common setup.}
All learned models are trained offline on trajectories generated from training instances using standard CG (unstabilized for CSP, and Du Merle stabilized for GAP). We select hyperparameters using validation instances and keep them fixed during test evaluation. Our primary efficiency metric is the \textit{paired runtime ratio}, computed by dividing each method's runtime by standard CG's runtime on the same instance and then averaging these ratios across instances. This normalizes for differences in instance size and difficulty, and gives each test instance equal weight. We additionally report the mean number of non-initial columns generated before convergence and the mean wall-clock runtime, averaged across instances. Results report the mean and standard error over independent test instances. Further training, hyperparameter, and implementation details are given in Appendix~\ref{app:learning_details}.

\subsection{Cutting Stock Problem}
\label{sec:experiments_csp}

\paragraph{Setup.}
We generate CSP instances using the CUTGEN1 generator of \citet{gau1995cutgen1}. We fix the stock length to \(L=10\,000\) and set the average demand to 10. Following \citet{kraul2023machine}, for each instance we sample the lower and upper relative item-length bounds independently as \(v_1\sim U[0.05,0.45]\) and \(v_2\sim U[0.50,0.85]\). We consider larger instances, sampling the number of item types \(n\) uniformly from \(\{500,\ldots,1500\}\), compared with \(\{50,\ldots,100\}\) in \citet{kraul2023machine}. We then sample \(n\) distinct integer item lengths uniformly without replacement from \([\lceil v_1L\rceil,\lfloor v_2L\rfloor]\). If this interval contains fewer than \(n\) distinct lengths, we resample \(v_1\) and \(v_2\) until the condition is satisfied. We use 100 instances for training, 50 for validation, and a separate 50 for testing. Hyperparameters are selected by mean paired runtime ratio relative to standard CG on the validation set. For predictive smoothing, this selects \(k=50\), \(\alpha_0=1\), and \(\gamma=0.9\).

\paragraph{Baselines.}
We compare against standard unstabilized CG, which prices using the current RMP duals. We then consider two classical stabilization strategies. Du Merle stabilization \citep{du1999stabilized} represents box-based stabilization with penalties, while Neame smoothing \citep{neame2000nonsmooth} represents dual smoothing by using a convex combination of the current dual and the previous smoothed pricing vector. Finally, we include the learned stabilization method of \citet{kraul2023machine}, which was developed specifically for the CSP. This method predicts the terminal dual once from static instance features and uses this fixed prediction as the center of a Du Merle stabilized master.

\paragraph{Comparison with existing methods.}

We report test set results in Table~\ref{tab:csp_kraul_scale_mixed}. Predictive smoothing achieves the lowest paired runtime ratio, reducing this by 37\% relative to standard CG on average. Neame smoothing gives a smaller improvement, with a paired runtime ratio of 0.865, while \citet{kraul2023machine} remains close to standard CG at 0.987.\footnote{The \citep{kraul2023machine} result is weaker than reported in their original study, under our setting with less training data and larger instances. We verified that our implementation reproduces their results under their original setting.} Du Merle stabilization is slower than standard CG. Predictive smoothing also generates fewer columns than all baselines, with reductions of 27\% relative to standard CG and 20\% relative to \citet{kraul2023machine}.

\begin{table}
    \caption{
        CSP results. Reported values are means and standard errors of per-instance metrics.
    }
    \label{tab:csp_kraul_scale_mixed}
    \begin{center}
    \begin{tabular}{lccc}
        \toprule
        Method
        & Paired runtime ratio $\downarrow$
        & Columns added $\downarrow$
        & Runtime (s) $\downarrow$ \\
        \midrule

        Standard CG
            & 1.000 \pmpar{0.000}
            & 2394 \pmpar{97}
            & 33.49 \pmpar{3.11} \\

        Du Merle stabilization
            & 1.266 \pmpar{0.019}
            & 2500 \pmpar{93}
            & 43.21 \pmpar{4.30} \\

        Neame smoothing
            & 0.865 \pmpar{0.005}
            & 2119 \pmpar{92}
            & 30.22 \pmpar{2.88} \\

        \citet{kraul2023machine}
            & 0.987 \pmpar{0.036}
            & 2207 \pmpar{93}
            & 29.24 \pmpar{2.70} \\

        Predictive smoothing (Ours)
            & \textbf{0.630} \pmpar{0.016}
            & \textbf{1756} \pmpar{102}
            & \textbf{24.84} \pmpar{2.70} \\

        \bottomrule
    \end{tabular}
    \end{center}
\end{table}

\begin{figure*}
    \centering
    \includegraphics[width=0.9\textwidth]{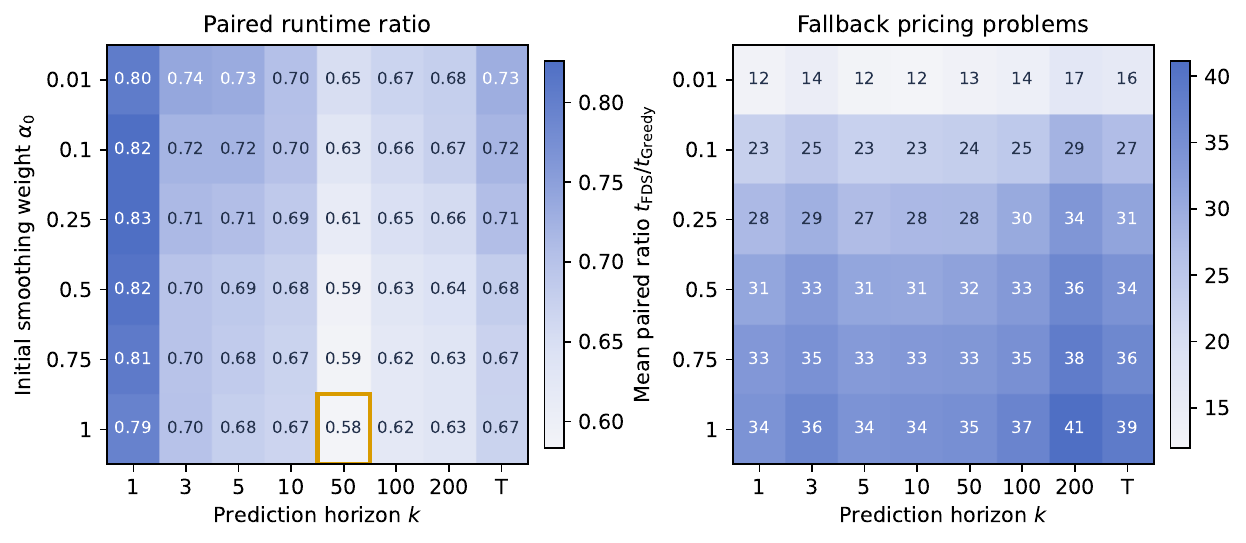}
    \caption{
        Validation performance of predictive dual smoothing as a function of horizon $k$ and initial smoothing weight $\alpha_0$. The best configuration is marked in orange.
    }
    \label{fig:csp-k-alpha-sweep}
\end{figure*}

\paragraph{Effect of the prediction horizon and smoothing strength.}
Figure~\ref{fig:csp-k-alpha-sweep} shows the validation results across prediction horizons $k$ and initial smoothing weights $\alpha_0$. Predictive smoothing performs best at an intermediate horizon, with $k=50$ achieving the lowest paired runtime ratio, while both $k=1$ and prediction of the terminal dual are less effective. Several factors likely contribute to this trade-off. Very short horizons provide limited look-ahead, so pricing remains sensitive to short-term fluctuations, whereas very long horizons may become too detached from the needs of the current RMP. Prediction accuracy also decreases with $k$ (Appendix~\ref{app:predictive_accuracy}), and larger horizons trigger somewhat more fallback pricing problems, as shown in the right panel of Figure~\ref{fig:csp-k-alpha-sweep}.

From Figure~\ref{fig:csp-k-alpha-sweep}, we also observe that a small smoothing weight already provides a significant benefit. At $\alpha_0=0.01$, predictive smoothing consistently improves over standard pricing. This is because the CSP's pricing problem often has many optimal solutions with identical reduced cost. Because of this, a small perturbation towards the predicted future dual already acts as an informed tie-breaker among these columns.

Finally, we observe that the best paired runtime ratio is obtained at $\alpha_0=1$. Although strong initial smoothing increases the number of fallback pricing problems, the corresponding reduction in CG iterations more than compensates for this additional work. In Appendix~\ref{app:gamma_ablation}, we show that it is the smoothing strength decay mechanism that allows strong initial smoothing $\alpha_0 = 1$ to work well. Also note that $\alpha_0=1$ and $\gamma=0.9$ are also selected in tuning for GAP. We thus find that these are robust default values to use. 

\paragraph{Training data efficiency.}
Table~\ref{tab:data_efficiency} shows that predictive smoothing degrades gradually as the amount of training data is reduced. Performance remains strong with as few as five training instances, while training on a single instance leads to overfitting and poor validation accuracy, resulting in worse CG performance. The comparison with \citet{kraul2023machine} also shows that predictive smoothing benefits more from additional training data: both validation accuracy and runtime improve substantially as $N$ increases, whereas Kraul's downstream performance is largely unaffected.

\begin{table}
\caption{Training data efficiency results for the CSP.}
\label{tab:data_efficiency}
\begin{center}
\begin{tabular}{r c c c c c c}
\toprule
& \multicolumn{3}{c}{Predictive Smoothing (Ours)} & \multicolumn{3}{c}{\citep{kraul2023machine}} \\
\cmidrule(lr){2-4} \cmidrule(lr){5-7}
$N$
& Validation $R^2$ $\uparrow$ & Paired ratio $\downarrow$ & Cols. $\downarrow$
& Validation $R^2$ $\uparrow$ & Paired ratio $\downarrow$ & Cols. $\downarrow$ \\
\midrule
1  & -0.858 & 1.559 & 3211 & 0.017 & 1.222 & 2507 \\
5  &  0.836 & 0.744 & 1904 & 0.424 & 0.920 & 2193 \\
10 &  0.857 & 0.712 & 1853 & 0.700 & 0.920 & 2176 \\
20 &  0.869 & 0.703 & 1852 & 0.738 & 0.930 & 2150 \\
50 &  0.873 & 0.678 & 1823 & 0.765 & 0.935 & 2114 \\
100 & 0.873 & 0.630 & 1756 & 0.782 & 0.987 & 2207 \\
\bottomrule
\end{tabular}
\end{center}
\end{table}

\paragraph{Out-of-distribution generalization.}

Table~\ref{tab:paired_runtime_ratios} shows that when generalizing to smaller instances of size $n=250$, predictive smoothing remains faster
than standard CG but is outperformed by Neame smoothing. A likely reason is that $k=50$ spans a much larger fraction of the shorter CG trajectories at this scale, making the predictor behave like a terminal predictor. In contrast, the method generalizes well to instances larger than those seen during training. At $n=2000$ and $n=2500$, it achieves paired runtime ratios of 0.646 and 0.695,
respectively, substantially outperforming all baselines.

\begin{table}
\caption{
Paired runtime ratios relative to standard CG on out-of-distribution test sets.
}
\label{tab:paired_runtime_ratios}
\begin{center}
\begin{tabular}{l c c c}
\toprule
& \multicolumn{3}{c}{Paired runtime ratio $\downarrow$} \\
\cmidrule(lr){2-4}

& \multicolumn{1}{c}{Smaller than training}
& \multicolumn{2}{c}{Larger than training} \\
\cmidrule(lr){2-2}
\cmidrule(lr){3-4}

Method
& $n=250$
& $n=2000$
& $n=2500$ \\
\midrule

Standard CG
& 1.000 \pmpar{0.000}
& 1.000 \pmpar{0.000}
& 1.000 \pmpar{0.000} \\

Neame smoothing
& \textbf{0.849} \pmpar{0.018}
& 0.843 \pmpar{0.018}
& 0.902 \pmpar{0.014} \\

Du Merle stabilization
& 1.253 \pmpar{0.035}
& 1.187 \pmpar{0.036}
& 1.312 \pmpar{0.044} \\

\citet{kraul2023machine}
& 1.187 \pmpar{0.073}
& 0.853 \pmpar{0.054}
& 0.895 \pmpar{0.057} \\

\textbf{Predictive smoothing (Ours)}
& 0.926 \pmpar{0.035}
& \textbf{0.646} \pmpar{0.036}
& \textbf{0.695} \pmpar{0.032} \\

\bottomrule
\end{tabular}
\end{center}
\end{table}

\subsection{Generalized Assignment Problem}

\paragraph{Setup.} The GAP involves assigning jobs to capacitated machines, where assignment costs and resource consumptions are specific to each job-machine combination. When addressed with CG, each machine leads to a separate pricing problem. At every CG iteration, we solve all machine pricing problems and add every optimal pattern with negative reduced cost to the RMP. Predictive smoothing is applied to the shared job-assignment duals. We use Martello-Toth Type C instances with $400$ jobs and $20$ machines~\citep{romeijn2001generating}. Training, validation and testing use $100$, $50$ and $50$ instances, respectively. The full formulation and implementation details are given in Appendix~\ref{app:gap}.
For predictive smoothing, tuning selected \(k=200\), \(\alpha_0=1\), and \(\gamma=0.9\).

\paragraph{Baselines.}
On GAP, Du Merle stabilization strongly outperforms standard unstabilized CG. We therefore evaluate predictive
smoothing on top of this stabilized master and compare against both Du Merle and Du Merle combined with Neame smoothing.

\paragraph{Comparison with existing methods.}
Du Merle stabilization reduces the paired runtime ratio to $0.078$, while adding Neame smoothing provides essentially no further improvement ($0.077$). Predictive smoothing instead reduces the ratio to $0.041$ and lowers the number of generated columns from 3327 to 2598. This shows how the two methods play complementary roles. Du Merle stabilization dampens dual oscillations, while predictive smoothing additionally steers pricing towards columns that address needs expected to arise in later CG iterations.

\begin{table}
\caption{GAP results. Reported values are means and standard errors of per-instance metrics.}
\label{tab:gap_results}
\begin{center}
\begin{tabular}{lccc}
    \toprule
    Method
    & Paired runtime ratio $\downarrow$
    & Columns added $\downarrow$
    & Runtime (s) $\downarrow$ \\
    \midrule

    Standard CG
        & 1.000 \pmpar{0.000}
        & 18991 \pmpar{104}
        & 388.93 \pmpar{5.55} \\

    Du Merle stabilization
        & 0.078 \pmpar{0.001}
        & 3327 \pmpar{12}
        & 30.42 \pmpar{0.43} \\

    Du Merle + Neame smoothing
        & 0.077 \pmpar{0.001}
        & 3063 \pmpar{10}
        & 29.74 \pmpar{0.44} \\

    Du Merle + Pred. smooth. (Ours)
        & \textbf{0.041} \pmpar{0.001}
        & \textbf{2598} \pmpar{27}
        & \textbf{15.87} \pmpar{0.27} \\

    \bottomrule
\end{tabular}
\end{center}
\end{table}

\section{Conclusions and Future Work}

We introduced \emph{predictive dual smoothing}, which stabilizes CG by smoothing the current pricing duals towards a learned prediction of a future dual state. The predictor is trained offline from standard CG trajectories, while reduced-cost checks and fallback pricing preserve correctness. Across cutting stock and generalized assignment, predictive smoothing reduces both generated columns and runtime, including when applied on top of strong master-side stabilization.
Future work includes embedding predictive smoothing within \emph{branch-and-price} schemes for integer linear programs, where column generation is repeatedly solved throughout the branch-and-bound tree. Another direction is to adapt the prediction horizon and smoothing strength online, potentially using measures of prediction uncertainty. Finally, evaluating transfer across instance distributions and CG formulations would clarify how broadly the approach generalizes.

\bibliography{preprint}

@article{chi2022deep,
  title={A deep reinforcement learning framework for column generation},
  author={Chi, Cheng and Aboussalah, Amine and Khalil, Elias and Wang, Juyoung and Sherkat-Masoumi, Zoha},
  journal={Advances in Neural Information Processing Systems},
  volume={35},
  pages={9633--9644},
  year={2022}
}

@article{morabit2021machine,
  title={Machine-learning--based column selection for column generation},
  author={Morabit, Mouad and Desaulniers, Guy and Lodi, Andrea},
  journal={Transportation Science},
  volume={55},
  number={4},
  pages={815--831},
  year={2021},
  publisher={INFORMS}
}

@inproceedings{fang2023accelerating,
  title={Accelerating Column Generation Algorithm Using Machine-Learning-Based Column Elimination},
  author={Fang, Lichang and Yuan, Haofeng and Zhang, Yuli and Song, Shiji},
  booktitle={2023 IEEE International Conference on Systems, Man, and Cybernetics (SMC)},
  pages={1945--1950},
  year={2023},
  organization={IEEE}
}

@inproceedings{multiple_columns,
  title={A reinforcement-learning-based multiple-column selection strategy for column generation},
  author={Yuan, Haofeng and Fang, Lichang and Song, Shiji},
  booktitle={Proceedings of the AAAI Conference on Artificial Intelligence},
  volume={38},
  number={8},
  pages={8209--8216},
  year={2024}
}

@article{ml_branch_and_price,
  title={A machine learning approach to rank pricing problems in branch-and-price},
  author={Kouteck{\'a}, Pavl{\'\i}na and {\v{S}}{\r{u}}cha, P{\v{r}}emysl and H{\r{u}}la, Jan and Maenhout, Broos},
  journal={European Journal of Operational Research},
  volume={320},
  number={2},
  pages={328--342},
  year={2025},
  publisher={Elsevier}
}

@article{ML_arc_selection,
  title={Machine-learning--based arc selection for constrained shortest path problems in column generation},
  author={Morabit, Mouad and Desaulniers, Guy and Lodi, Andrea},
  journal={INFORMS Journal on Optimization},
  volume={5},
  number={2},
  pages={191--210},
  year={2023},
  publisher={INFORMS}
}

@inproceedings{hu2025ffcg,
  title={FFCG: Effective and Fast Family Column Generation for Solving Large-Scale Linear Program},
  author={Hu, Yi-Xiang and Wu, Feng and Li, Shaoang and Zhao, Yifang and Li, Xiang-Yang},
  booktitle={Proceedings of the AAAI Conference on Artificial Intelligence},
  volume={39},
  number={11},
  pages={11238--11245},
  year={2025}
}

@article{kraul2023machine,
  title={Machine learning--supported prediction of dual variables for the cutting stock problem with an application in stabilized column generation},
  author={Kraul, Sebastian and Seizinger, Markus and Brunner, Jens O},
  journal={INFORMS Journal on Computing},
  volume={35},
  number={3},
  pages={692--709},
  year={2023},
  publisher={INFORMS}
}

@book{babaki2022coil,
  title={COIL: A deep architecture for column generation},
  author={Babaki, Behrouz and Charlin, Laurent and Jena, Sanjay Dominik},
  year={2022},
  publisher={Bureau de Montreal, Universit{\'e} de Montreal}
}

@inproceedings{fang2025learning,
  title={Learning to Stabilize Column Generation},
  author={Fang, Lichang and Yuan, Haofeng and Song, Shiji and Chen, Bokui},
  booktitle={2025 International Joint Conference on Neural Networks (IJCNN)},
  pages={1--8},
  year={2025},
  organization={IEEE}
}

@inproceedings{shen2022pricing,
  title={Enhancing column generation by a machine-learning-based pricing heuristic for graph coloring},
  author={Shen, Yunzhuang and Sun, Yuan and Li, Xiaodong and Eberhard, Andrew and Ernst, Andreas},
  booktitle={Proceedings of the AAAI conference on artificial intelligence},
  volume={36},
  number={9},
  pages={9926--9934},
  year={2022}
}

@inproceedings{sun2022learning,
  title={Learning to generate columns with application to vertex coloring},
  author={Sun, Yuan and Ernst, Andreas T and Li, Xiaodong and Weiner, Jake},
  booktitle={The Eleventh International Conference on Learning Representations},
  year={2022}
}

@article{shen2024adaptive,
  title={Adaptive stabilization based on machine learning for column generation},
  author={Shen, Yunzhuang and Sun, Yuan and Li, Xiaodong and Cao, Zhiguang and Eberhard, Andrew and Zhang, Guangquan},
  journal={arXiv preprint arXiv:2405.11198},
  year={2024}
}

@phdthesis{neame2000nonsmooth,
  title={Nonsmooth dual methods in integer programming},
  author={Neame, Philip James},
  year={2000},
  school={University of Melbourne, Department of Mathematics and Statistics}
}

@article{du1999stabilized,
  title={Stabilized column generation},
  author={Du Merle, Olivier and Villeneuve, Daniel and Desrosiers, Jacques and Hansen, Pierre},
  journal={Discrete Mathematics},
  volume={194},
  number={1-3},
  pages={229--237},
  year={1999},
  publisher={Elsevier}
}

@article{gau1995cutgen1,
  title={CUTGEN1: A problem generator for the standard one-dimensional cutting stock problem},
  author={Gau, T and W{\"a}scher, G},
  journal={European journal of operational research},
  volume={84},
  number={3},
  pages={572--579},
  year={1995},
  publisher={Elsevier}
}

@article{gilmore1961linear,
  title={A linear programming approach to the cutting-stock problem},
  author={Gilmore, Paul C and Gomory, Ralph E},
  journal={Operations research},
  volume={9},
  number={6},
  pages={849--859},
  year={1961},
  publisher={INFORMS}
}

@article{wentges1997weighted,
  title={Weighted Dantzig-Wolfe decomposition for linear mixed-integer programming},
  author={Wentges, Paul},
  journal={International Transactions in Operational Research},
  volume={4},
  number={2},
  pages={151--162},
  year={1997},
  publisher={Elsevier}
}

@article{romeijn2001generating,
  title={Generating experimental data for the generalized assignment problem},
  author={Romeijn, H Edwin and Romero Morales, Dolores},
  journal={Operations Research},
  volume={49},
  number={6},
  pages={866--878},
  year={2001},
  publisher={INFORMS}
}

@article{sarin2014branch,
  title={A branch-and-price approach for the stochastic generalized assignment problem},
  author={Sarin, Subhash C and Sherali, Hanif D and Kim, Seon Ki},
  journal={Naval Research Logistics (NRL)},
  volume={61},
  number={2},
  pages={131--143},
  year={2014},
  publisher={Wiley Online Library}
}

@article{savelsbergh1997branch,
  title={A branch-and-price algorithm for the generalized assignment problem},
  author={Savelsbergh, Martin},
  journal={Operations research},
  volume={45},
  number={6},
  pages={831--841},
  year={1997},
  publisher={INFORMS}
}

@article{lubbecke2005selected,
  title={Selected topics in column generation},
  author={L{\"u}bbecke, Marco E and Desrosiers, Jacques},
  journal={Operations research},
  volume={53},
  number={6},
  pages={1007--1023},
  year={2005},
  publisher={INFORMS}
}

@book{desaulniers2006column,
  title={Column generation},
  author={Desaulniers, Guy and Desrosiers, Jacques and Solomon, Marius M},
  year={2006},
  publisher={Springer Science \& Business Media}
}

@incollection{desrosiers2005primer,
  title={A primer in column generation},
  author={Desrosiers, Jacques and L{\"u}bbecke, Marco E},
  booktitle={Column generation},
  pages={1--32},
  year={2005},
  publisher={Springer}
}

@article{pessoa2018automation,
  title={Automation and combination of linear-programming based stabilization techniques in column generation},
  author={Pessoa, Artur and Sadykov, Ruslan and Uchoa, Eduardo and Vanderbeck, Fran{\c{c}}ois},
  journal={INFORMS Journal on Computing},
  volume={30},
  number={2},
  pages={339--360},
  year={2018},
  publisher={INFORMS}
}
\bibliographystyle{iclr2027_conference}

\newpage
\appendix

\section{Feature Representation}
\label{app:features}

We construct the feature representation from quantities that are available directly from the current RMP and pricing formulation in state $s_t$. Consider an RMP in canonical form
\begin{equation*}
    \min_{\lambda \geq 0}
    \sum_{p\in\mathcal{P}_t} c_p \lambda_p
    \qquad
    \text{s.t.}
    \qquad
    \sum_{p\in\mathcal{P}_t} A_{ip}\lambda_p \geq b_i,
    \quad i=1,\ldots,n,
\end{equation*}
with optimal dual solution $\boldsymbol{\pi}_t$ in iteration $t$. For each master constraint $i$, we construct a fixed-dimensional feature vector $f_i(s_t)$. The representation uses only quantities derived from the CG iteration, the corresponding master row, the pricing formulation, and the columns currently present in the RMP. Concretely, we use the following four groups of features:

\begin{enumerate}
    \item \textbf{Global CG state.} We include the current iteration $t$, the number of columns $|\mathcal{P}_t|$ in the RMP, the current RMP objective value, and the mean reduced cost of the current columns.

    \item \textbf{Constraint state.} For constraint $i$, we include its current dual value $\pi_{i,t}$, right-hand side $b_i$, and current primal activity
    $a_{i,t} = \sum_{p\in\mathcal{P}_t} A_{ip}\lambda_p$.

    \item \textbf{Pricing structure.} We identify the pricing variables whose values determine the coefficient $A_{ip}$ of a newly generated column in row $i$. We include the minimum, mean, and maximum of their coefficients in the pricing objective and pricing constraints. When multiple variables or pricing subproblems are associated with the same master row, we use the minimum, mean, and maximum. Pricing-constraint coefficients are additionally normalized by their corresponding right-hand sides.
    
    \item \textbf{Column context.} The fourth group summarizes how constraint $i$ is represented by the columns already generated. Let $\mathcal{P}_t(i) = \{p\in\mathcal{P}_t : A_{ip}\neq 0\}$ denote the current columns that contain a nonzero coefficient in row $i$. We include the minimum positive, mean, and maximum values of $A_{ip}$ over the current column set, together with the fraction $|\mathcal{P}_t(i)|/|\mathcal{P}_t|$ of columns containing the row. We also include row-specific statistics of the reduced costs of columns in $\mathcal{P}_t(i)$, including the fraction whose reduced cost is close to zero. Finally, when pricing decomposes into multiple blocks or subproblems, we include summary statistics of the pricing-block quantities associated with the columns in $\mathcal{P}_t(i)$.
\end{enumerate}

Features are standardized using training-set statistics. The same transformations are applied at validation and test time.

\section{Learning and Experimental Details}
\label{app:learning_details}

\subsection{Predictor Architecture}

We train a separate predictor for each horizon $k\in\{1,3,5,10,50,100,200,\text{terminal}\}$. Each predictor is a row-wise MLP that maps the fixed-dimensional feature vector $f_i(s_t)$ of one RMP constraint to a scalar future-dual prediction. The same MLP is applied independently to every RMP constriant, so its parameter count is independent of instance size.

For both CSP and GAP, the MLP has one fully-connected hidden layer with 32 ReLU units and a linear output layer. Input features are standardized, but target dual values are not, so predictions are produced directly on the original dual scale.

For CSP, the target at state $s_t$ for horizon $k$ is $\boldsymbol{\pi}_{\min(t+k,T)}$, while the terminal model targets $\boldsymbol{\pi}_T$ at every state. Labels are obtained from standard CG trajectories. GAP uses the same target convention, but predicts only the job-partitioning duals (i.e., the machine convexity duals are left unchanged, as they do not affect pricing). GAP labels are obtained from Du Merle trajectories with $\epsilon=0.5$.

RMP dual solutions are not always unique. We experimented with always using the minimum-norm duals of the optimal dual face as canonical targets (which requires solving a QP per state at data collection time), but found no improvement in downstream CG performance, so we use the duals returned by the LP solver.

\subsection{Training Procedure}

Training, validation, and test instances are generated using disjoint random seeds. For both problem classes, we use 100 training instances, 50 validation instances, and 50 test instances.

For CSP, we first collect complete standard CG trajectories. Retaining every state and row would produce an unnecessarily large training set, so we subsample each trajectory to obtain data from diverse instances, while keeping training manageable. Concretely, from each training instance we sample 50 states stratified over the trajectory, choosing one state uniformly at random from each of 50 equal intervals. From each selected state, we then sample 256 item rows uniformly at random without replacement. We do this because we found that covering a larger and more diverse set of instances was more useful than using complete trajectories from fewer instances. Each instance therefore contributes $50\times256=12\,800$ constraint-level examples, leading to approximately $1.28$ million training examples. The validation set is constructed independently using the same procedure, producing approximately $640\,000$ validation examples. Sampled states and rows are shared across horizons, with only the future-dual targets changing with $k$.

For GAP, we retain all job rows from mature states of the Du Merle trajectories. A state is considered mature once no more artificial Big-$M$ initialization patterns are used anymore in optimal RMP solution. Earlier startup states are excluded from training and validation, and predictive smoothing is likewise disabled during this phase at test time.

Continuous features are standardized using means and standard deviations computed from training rows only, and the same transformation is reused for validation and test inference. Near-constant standard deviations are replaced by one for numerical stability.

All predictive smoothing predictors minimize mean squared error using Adam with learning rate $10^{-2}$. CSP uses mini-batches of 8192 examples; GAP uses mini-batches of 4096. Examples are shuffled at each epoch.

Training is capped at 100 epochs and uses validation MSE for early stopping. The used paatience is three validation checks, and patience is reset only when validation MSE improves by at least $0.5\%$ relative to the current best value. The checkpoint with the lowest validation MSE is restored before evaluation and saving.

\subsection{Hyperparameter Selection}

All hyperparameters are selected using the validation set. Validation instances may vary in size, so  configurations are ranked by mean paired runtime ratio, $(1/|\mathcal V|)\sum_{i\in\mathcal V} T_h(i)/T_{\mathrm{Standard}}(i)$. This weights each instance equally and avoids having larger/harder instances dominate the raw mean runtime.

For CSP, we evaluated predictive smoothing with horizons $k\in\{1,3,5,10,50,100,200,\text{terminal}\}$, initial smoothing weights $\alpha_0\in\{0.01,0.1,0.25,0.5,0.75,1\}$, and alpha decay factor $\gamma \in \{0.1, 0.5, 0.9, 1\}$. For Neame smoothing, we evaluated the same $\alpha$ grid. For Du Merle stabilization, we evaluated $\epsilon\in\{0.03125,0.0625,0.125,0.25,0.5,1,2,4, 8\}$. Validation selected $\alpha=0.75$ for Neame, $\epsilon=0.125$ for Du Merle, $\epsilon=2$ for Kraul, and $k=50$, $\alpha_0=1$, $\gamma=0.9$ for predictive smoothing.

For GAP, we evaluated the same grids. Validation selected $\epsilon = 0.5$ for Du Merle stabilization. When tuning smoothing methods used in combination with Du Merle stabilization, we tune the remaining hyperparameters for smoothing after setting the stabilization $\epsilon = 0.5$. Validation selected $\alpha=0.25$ for Du Merle with Neame smoothing, and $k=200$, $\alpha_0=1$, and $\gamma=0.9$ for Du Merle with predictive smoothing.

During mature GAP states, predictive smoothing is applied only to the job duals. If no machine produces an improving pattern under predictive pricing, all machine pricing problems are rerun using the exact current job duals, and $\alpha_t$ is decayed only after this global fallback. Du Merle's $\epsilon$ is halved only when neither predictive nor exact pricing produces a new column, and termination requires a final run of exact pricing.

\subsection{Implementation Details}
RMPs were solved using Gurobi 12.0.3 with one solver thread, while pricing used an exact Numba dynamic-programming implementation. Experiments were run on Ubuntu 24.04.4 LTS, using an AMD EPYC 9334 CPU and 252 GiB of RAM.

\section{Generalized Assignment Problem}
\label{app:gap}

The generalized assignment problem (GAP) involves assigning $n$ jobs to $m$ machines. Assigning job $j$ to machine $i$ incurs cost $c_{ij}$ and consumes $a_{ij}$ units of machine capacity, where machine $i$ has capacity $b_i$. Each job must be assigned to exactly one machine, and the objective is to minimize total assignment cost without exceeding any machine's capacity.

We use a pattern-based master formulation for GAP. Let $\mathcal P_i$ denote the set of feasible job subsets for machine $i$. A pattern $p\in\mathcal P_i$ has incidence vector $a^p\in\{0,1\}^n$, where $a_j^p=1$ if job $j$ is assigned to machine $i$, and cost $c_p=\sum_j c_{ij}a_j^p$. Feasibility requires $\sum_j a_{ij}a_j^p\leq b_i$.

At iteration $t$, the RMP contains subsets $\mathcal P_{i,t}\subseteq\mathcal P_i$ and a nonnegative variable $\lambda_p$ for each generated pattern:
\begin{align}
    \min_{\lambda\geq0}\quad
    & \sum_{i=1}^m\sum_{p\in\mathcal P_{i,t}} c_p\lambda_p \\
    \text{s.t.}\quad
    & \sum_{i=1}^m\sum_{p\in\mathcal P_{i,t}} a_j^p\lambda_p=1,
    && j=1,\ldots,n, \\
    & \sum_{p\in\mathcal P_{i,t}}\lambda_p\leq1,
    && i=1,\ldots,m.
\end{align}
The first constraints are job-partitioning constraints and the second are machine-convexity constraints. Let $\pi_j$ denote the unrestricted dual of job partitioning constraint $j$ and $\sigma_i\leq0$ the dual of the convexity constraint for machine $i$.

Pricing decomposes into one $0$-$1$ knapsack problem per machine:
\begin{align}
    \max_{x\in\{0,1\}^n}\quad&
        \sum_{j=1}^n (\pi_j-c_{ij})x_j + \sigma_i
        \\
    \text{s.t.}\quad&
        \sum_{j=1}^n a_{ij}x_j \leq b_i.
\end{align}
At each CG iteration, we solve the pricing problem for each machine, obtaining one optimal pattern per machine. We then add every pattern among these optima whose reduced cost is negative.

To obtain an initially feasible RMP, we use a Big-\(M\) artificial-column initialization similar to that used in branch-and-price approaches for GAP \citep{savelsbergh1997branch, sarin2014branch}. We introduce one artificial column for each job. Each artificial column covers only that job, and has objective cost of a large value $M$, which makes artificial columns unattractive once feasible genuine patterns are available. During CG, we consider this initialization phase complete once no more artificial columns are used in the optimal RMP solution. When collecting training data for GAP, we only collect data after the initialization phase. Similarly, at test time, predictive smoothing and Du Merle stabilization are disabled during this initialization phase.

We use Martello--Toth Type C instances with $a_{ij}\sim\mathrm{Uniform}\{5,\ldots,25\}$ and independently $c_{ij}\sim\mathrm{Uniform}\{10,\ldots,50\}$, with $b_i=0.8\sum_j a_{ij}/m$ \citep{romeijn2001generating}. We use $n=400$ jobs and $m=20$ machines.

\section{Predictive accuracy on CSP}
\label{app:predictive_accuracy}
Table~\ref{tab:fds_validation_metrics} reports the predictive accuracy of the future-dual predictor on the CSP validation set for different prediction horizons $k$. Prediction becomes progressively more difficult as the horizon increases. Validation RMSE generally increases and $R^2$ decreases as the target dual lies further along the CG trajectory.

\begin{table}[h!]
\caption{Predictive accuracy of the predictor as a function of the horizon $k$.}
\label{tab:fds_validation_metrics}
\begin{center}
\begin{tabular}{lcccccccc}
\toprule
Prediction horizon $k$
    & 1 & 3 & 5 & 10 & 50 & 100 & 200 & Terminal \\
\midrule
Validation RMSE
    & 0.078 & 0.091 & 0.090 & 0.097
    & 0.112 & 0.119 & 0.124 & 0.131 \\
Validation $R^2$
    & 0.938 & 0.916 & 0.918 & 0.906
    & 0.873 & 0.855 & 0.838 & 0.802 \\
\bottomrule
\end{tabular}
\end{center}
\end{table}

\section{Ablation of smoothing strength decay $\gamma$}
\label{app:gamma_ablation}

\begin{figure}
    \centering
    \includegraphics[width=0.6\linewidth]{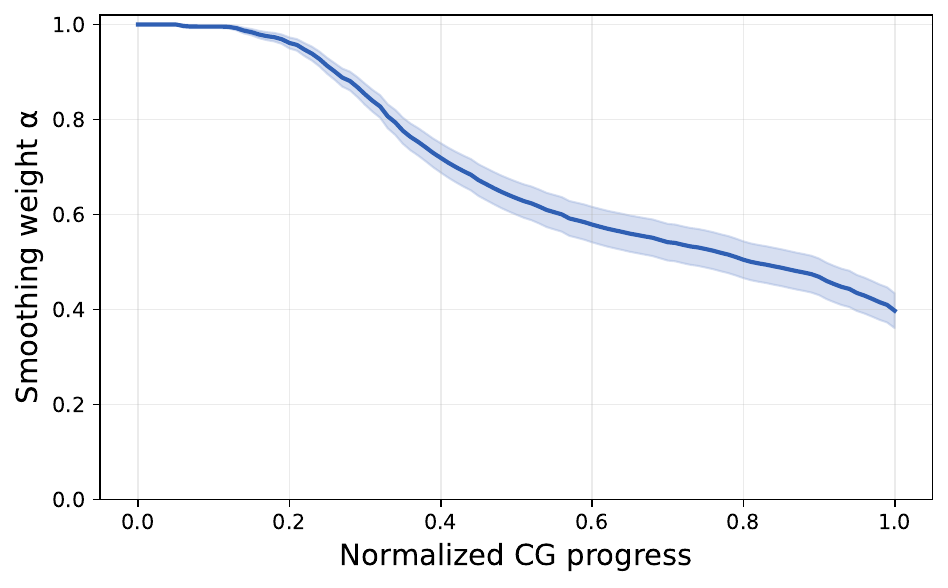}
    \caption{Evolution of the smoothing strength $\alpha_t$ over normalized CG progress for the selected CSP configuration ($k=50$, $\alpha_0=1$, $\gamma=0.9$). The line shows the mean across validation instances and the shaded region shows SEM across instances.}
    \label{fig:alpha_evolution}
\end{figure}

Figure~\ref{fig:alpha_evolution} shows how the smoothing strength $\alpha_t$ evolves for the selected CSP configuration ($k=50$, $\alpha_0=1$, $\gamma=0.9$). The method retains strong predictive smoothing early in CG, while fallbacks progressively reduce $\alpha_t$ later in the trajectory, gradually shifting pricing towards the current RMP dual.

\begin{figure}
    \centering
    \begin{subfigure}{0.8\linewidth}
        \centering
        \includegraphics[width=\linewidth]{fds_adaptive_fallback_paired_heatmaps.pdf}
        \caption{$\gamma=0.9$}
        \label{fig:gamma_ablation_0_9}
    \end{subfigure}

    \begin{subfigure}{0.8\linewidth}
        \centering
        \includegraphics[width=\linewidth]{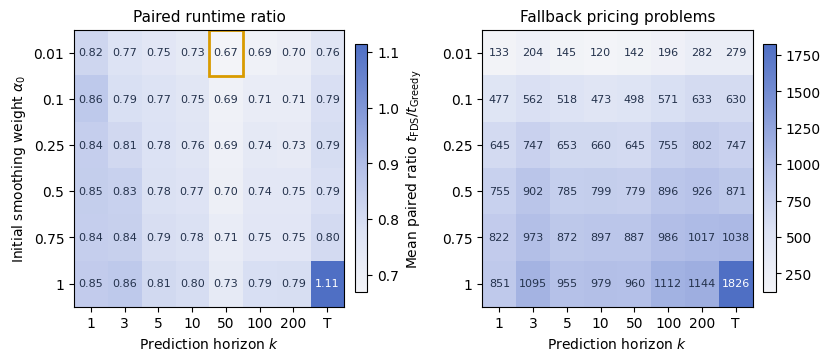}
        \caption{$\gamma=1$ (no decay)}
        \label{fig:gamma_ablation_1}
    \end{subfigure}

    \caption{Effect of smoothing decay $\gamma$ on CSP validation performance. Each panel reports the paired runtime ratio and number of fallback pricing problems across prediction horizons $k$ and initial smoothing weights $\alpha_0$.}
    \label{fig:gamma_ablation}
\end{figure}

This decay is important for efficiency.
Figures~\ref{fig:gamma_ablation_0_9} and~\ref{fig:gamma_ablation_1} compare predictive smoothing with $\gamma=0.9$ against the non-decaying case $\gamma=1$. Decaying the smoothing strength after a fallback significantly reduces both runtime and the number of fallback pricing problems. $\gamma=0.9$ consistently improves runtime across the grid and makes performance considerably less sensitive to large $\alpha_0$. This supports the adaptive rule used in our main experiments. Predictive smoothing can remain strong while its guidance is useful, and repeated fallbacks provides a simple signal to move pricing progressively closer to the current RMP dual as CG approaches convergence.

\end{document}